\documentclass[conference]{IEEEtran}
\IEEEoverridecommandlockouts

\usepackage{cite}
\usepackage{amsmath,amssymb,amsfonts}
\usepackage{algorithmic}
\usepackage{graphicx}
\usepackage{textcomp}
\usepackage{multirow}
\usepackage{xcolor}
\def\BibTeX{{\rm B\kern-.05em{\sc i\kern-.025em b}\kern-.08em
    T\kern-.1667em\lower.7ex\hbox{E}\kern-.125emX}}
\begin{document}

\title{Emotion Detection on User Front-Facing App Interfaces for Enhanced Schedule Optimization: A Machine Learning Approach
\thanks{Accepted in IISA 2025 and pending publication in IEEE.}
}

\author{
\IEEEauthorblockN{Feiting Yang \IEEEauthorrefmark{1}, Antoine Moevus \IEEEauthorrefmark{2}, Steve Lévesque \IEEEauthorrefmark{3}}
\IEEEauthorblockA{\IEEEauthorrefmark{1} The Edward S. Rogers Sr. Department of Electrical and Computer Engineering, University of Toronto, Toronto, Canada \\
Email: {feiting.yang@mail.utoronto.ca}}
\IEEEauthorblockA{\IEEEauthorrefmark{2} Département d'informatique et de recherche opérationnelle (DIRO), Université de Montréal, Montreal, Canada \\
Email: {antoine.moevus@umontreal.ca}}
\IEEEauthorblockA{\IEEEauthorrefmark{3} Department of Software and IT Engineering, École de technologie supérieure, Montreal, Canada \\
Email: {steve.levesque.1@ens.etsmtl.ca}}
}

\maketitle

\begin{abstract}

Human-Computer Interaction (HCI) has evolved significantly to incorporate emotion recognition capabilities, creating unprecedented opportunities for adaptive and personalized user experiences. This paper explores the integration of emotion detection into calendar applications, enabling user interfaces to dynamically respond to users' emotional states and stress levels, thereby enhancing both productivity and engagement. We present and evaluate two complementary approaches to emotion detection: a biometric-based method utilizing heart rate (HR) data extracted from electrocardiogram (ECG) signals processed through Long Short-Term Memory (LSTM) and Gated Recurrent Unit (GRU) neural networks to predict the emotional dimensions of Valence, Arousal, and Dominance; and a behavioral method analyzing computer activity through multiple machine learning models to classify emotions based on fine-grained user interactions such as mouse movements, clicks, and keystroke patterns. Our comparative analysis, from real-world datasets, reveals that while both approaches demonstrate effectiveness, the computer activity-based method delivers superior consistency and accuracy, particularly for mouse-related interactions, which achieved approximately 90\% accuracy. Furthermore, GRU networks outperformed LSTM models in the biometric approach, with Valence prediction reaching 84.38\% accuracy.
\end{abstract}

\begin{IEEEkeywords}
Machine Learning, Deep Learning, UI/UX, HCI, Calendar Optimization, Computer Activity, Emotion Detection
\end{IEEEkeywords}

\section{Introduction}  

Human-Computer Interaction (HCI) has traditionally focused on enhancing user experiences by creating systems that are functional, intuitive, and engaging. However, conventional digital productivity tools often neglect a critical dimension of user experience: the emotional state of the user. In recent years, the integration of emotion recognition into HCI has introduced new opportunities for developing truly personalized interactions that adapt to users' emotional states, addressing a significant gap in current interface design \cite{HCI_emo}.

The growing challenge of information overload and schedule management in digital environments demands more sophisticated approaches to interface design. Studies indicate that knowledge workers increasingly struggle with information overload, which negatively impacts their cognitive performance and well-being \cite{arnold2023dealing, bubb2012information, soucek2010coping}. This problem is compounded by ineffective scheduling and time management systems. A survey of 182 senior managers across various industries found that 71\% considered meetings unproductive and inefficient, while 65\% reported that meetings prevented them from completing their own work \cite{perlow2017stop}. Despite these challenges, current calendar and scheduling applications remain emotionally unaware, failing to consider how stress levels and emotional states significantly affect productivity and decision-making \cite{fellmann2019stress}. This gap between user needs and system capabilities leads to suboptimal experiences and decreased efficiency in time management.

Emotion recognition in HCI involves identifying a user's emotional state through three primary signal categories: physiological, behavioral, and contextual. Physiological techniques analyze biometric data such as heart rate, facial expressions, and EEG signals to predict distinct emotional states characterized by dimensions of Valence (positive/negative), Arousal (intensity), and Dominance (sense of control) \cite{FacialDL,hans2021cnn,Deepher,choppin2000eeg}. Behavioral cues from device interactions, including mouse movements, clicks, and keystrokes, can detect shifts in emotions such as stress or relaxation \cite{Com_stress, akolakowska2013}. Contextual signals derived from user input text or speech provide additional emotional insights \cite{bharti2022text,nandwani2021review}. By interpreting these multidimensional emotional signals, systems can dynamically adjust to meet users' evolving needs.

Schedule management through a graphical calendar represents a particularly compelling application domain for emotion-aware HCI, offering significant potential to enhance both user experience and productivity. 
Throughout this paper, we will use the term "calendar optimization" to refer to schedule optimization through a graphical dynamic calendar.  
For this work we focus on an emotion-aware calendar system that can assess a user's emotional state through computer activity patterns and/or physiological data. This system then performs calendar optimization at two levels: i) interface adaptation—modifying visual elements and interaction patterns based on detected emotions, and ii) scheduling intelligence—applying constraint-based algorithms to generate emotionally-optimized event arrangements. For instance, detecting elevated stress levels might trigger interface simplification and suggest scheduling breaks to mitigate cognitive load, while positive emotional states could prompt recommendations for more challenging tasks when the user is best equipped to handle them.

The integration of emotion recognition with calendar optimization can be further enhanced through Constraint Satisfaction Problem (CSP) algorithms, which excel at finding optimal solutions within predefined constraints \cite{csp}. When augmented with emotional awareness, these algorithms can dynamically adjust scheduling by considering not only traditional constraints like time and resources, but also the user's current and predicted emotional states. This creates a uniquely personalized and adaptive experience that aligns tasks with emotional readiness, potentially transforming how users interact with time management tools.

This paper explores two complementary approaches for emotion detection specifically suited for calendar optimization applications. Our first approach leverages biometric signals, extracting HR data from ECG signals to predict the emotional dimensions of Valence, Arousal, and Dominance by using LSTM networks and GRU networks. Our second approach analyses computer activity patterns, employing multiple machine learning models including Random Forest, Support Vector Machines, and ensemble methods to classify emotions based on fine-grained user interactions. By combining these approaches, we address the limitations of single-modality emotion detection systems and provide a more robust foundation for emotion-aware interface design.

The innovation of our work lies in: i) the comparative analysis of biometric and behavioral approaches to emotion detection in a calendar optimization context; ii) the development of specific models for detecting emotional dimensions relevant to productivity; and iii) establishing an empirical foundation for emotion-driven UI/UX design that can significantly enhance scheduling systems.

The remainder of this paper is organized as follows: Section \ref{Sec:Architecture} describes the global emotion-based planification system, Section \ref{Sec:RelatedWork} provides an overview of the common approaches to emotion recognition, along with a detailed description of the deep learning models employed. Section \ref{Sec:Proposed} explores two novel approaches based on computer activity and heart rate signals. Section \ref{Sec:Experimental} explains the experimental results obtained. Finally, Section \ref{Sec:Conclusion} concludes the paper and outlines directions for future work.

\subsection{System Architecture}
\label{Sec:Architecture}

\begin{figure}[h]
    \centering
    \includegraphics[width=0.4\textwidth]{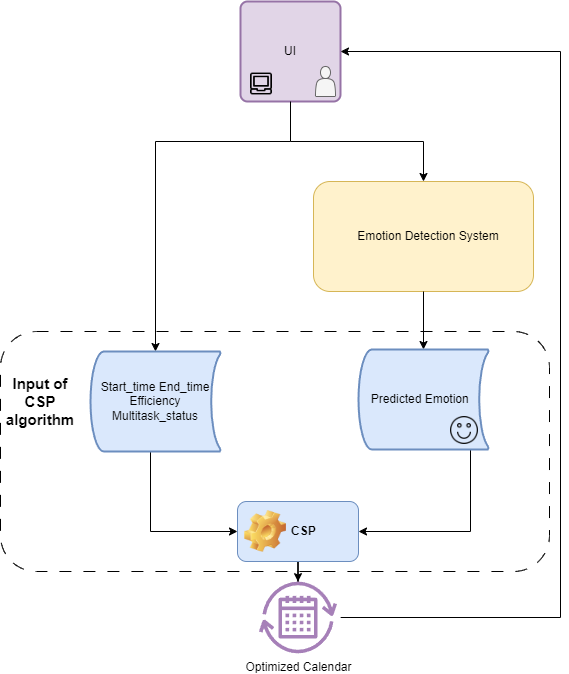}
    \caption{Overview of the generic system for emotion-based calendar optimization}
    \label{fig:system}
\end{figure}

The architecture of our emotion-based calendar optimization system is illustrated in Fig. \ref{fig:system}. This integrated system consists of three primary components working in concert. The first component is the a front-facing interface with which users interact. It serves both as an input mechanism and visualization layer. This interface simultaneously collects behavioral data while channeling it to the Emotion Detection System (EDS).

The second component is the EDS. It functions as the system's perceptual core, processing two distinct data streams: i) physiological signals captured from wearable devices such as heart rate data from fitness trackers or smartwatches, and ii) behavioral patterns extracted from computer interactions including mouse movements, keystroke dynamics, and click patterns. Advanced machine learning models within the EDS process these multimodal inputs to classify the user's current emotional state along the dimensions of Valence, Arousal, and Dominance.

The third component is an adaptive scheduling engine powered by a CSP solver. This component receives three input categories: the emotional state vector from the EDS, explicit user-defined parameters (start/end times, task priorities), and implicit contextual variables (efficiency metrics, multitasking capacity). The CSP algorithm processes these constraints through an optimization function that balances productivity requirements with emotional well-being considerations. The output is a dynamically generated calendar schedule that not only satisfies temporal constraints but also aligns task difficulty and cognitive demands with the user's detected emotional state, creating a personalized scheduling experience that adapts in real-time to changing emotional conditions.

\section{Related Work}
\label{Sec:RelatedWork}

This section examines the foundational techniques that inform our emotion-aware calendar optimization system. We first present constraint satisfaction algorithms for calendar scheduling, then survey the current landscape of emotion detection methodologies across multiple modalities. Finally, we review the neural network architectures employed in our approach.

\subsection{Constraint Satisfaction}
CSPs are a class of computational problems that involve finding an assignment of values to a set of variables such that a predefined set of constraints is satisfied. A CSP instance can be formally represented as a triplet \( (V, D, C) \), where:

\begin{itemize}
    \item \( V \) is a finite set of variables,
    \item \( D \) is a set of values, also known as the domain,
    \item \( C \) is a finite set of constraints \( \{ C_1, C_2, \dots, C_q \} \).
\end{itemize}

In emotion-aware calendar optimization, we extend the traditional CSP framework to incorporate emotional parameters that influence scheduling decisions.

The formulation of this extended problem is as follows:
\textbf{Variables and Parameters:}
Let \( E_1, E_2, \dots, E_n \) represent the events to be scheduled, where \( n \) is the total number of events. Each event \( E_i \) has the following properties:

\begin{itemize}
    \item \textbf{Temporal parameters:}
    \begin{itemize}
        \item Start time $s_i \in \mathbb{T}$, where $\mathbb{T}$ is the set of possible time slots
        \item Duration $d_i \in \mathbb{R}^+$, representing the length of the event
        \item End time $e_i = s_i + d_i$
    \end{itemize}
    
    \item \textbf{Efficiency parameters:}
    \begin{itemize}
        \item Multitasking capability $m_i \in \{0,1\}$, indicating whether the event can be performed concurrently with other events
        \item Priority coefficient $p_i \in [0,1]$, representing the relative importance of the event
    \end{itemize}
    
    \item \textbf{Emotional context:}
    \begin{itemize}
        \item Emotional state vector that represents the user emotions detected at scheduling time  $i$. E.g., $\vec{\varepsilon}_i = (\varepsilon_v, \varepsilon_a, \varepsilon_d)$, where $\varepsilon_v$, $\varepsilon_a$, and $\varepsilon_d$ represent the valence, arousal, and dominance components 
    \end{itemize}
\end{itemize}

\textbf{Domains:} Each event variable represents a set of feasible domains. For example:

\begin{itemize}
    \item The start time \( s_i \) typically belongs to a set of available time slots, such as \( \{ 9:00 \, \text{AM}, 9:30 \, \text{AM}, \dots, 6:00 \, \text{PM} \} \),
    \item The end time \( e_i \) is determined by the start time and duration, i.e., \( e_i = s_i + d_i \).
\end{itemize}

Thus, the domain for the start time can be expressed as:
\[
D(s_i) = \{ t \mid t \in [9:00 \, \text{AM}, 6:00 \, \text{PM}] \text{ and } e_i = s_i + d_i \}
\]

\textbf{Constraints Formulation:}

Several constraints govern the scheduling of events, including:
\begin{itemize}
    \item \textbf{Temporal Exclusivity Constraint:} Prevents temporal overlap between non-multitaskable events:
    \[
    \mathbf{C_1: s_i + d_i \leq s_j \quad \textbf{or} \quad s_j + d_j \leq s_i, \quad \forall i \neq j}
    \]

      \item \textbf{Priority-Based Sequencing Constraint:} Enforces scheduling high-priority events earlier when appropriate:

    \[
    \mathbf{C_2: s_i \quad \textbf{if} \quad p_i > p_j \, \textbf{or} \, m_i = \textbf{False}}
    \]

    \item \textbf{Emotional Regulation Constraint:} Ensures appropriate pacing based on detected emotional states:
    \[
    \mathbf{C_3}: \textbf{If Emotion}_i \geq \mathbf{T}_{\textbf{stress}},
    \]
    \[
    \quad \textbf{schedule low-demand tasks} 
    \]
    \[
    \quad \text{(e.g., relaxing activities or breaks) after } E_i
    \]
    where \( T_{\text{stress}} \) is a threshold indicating high stress or arousal.

    \item \textbf{Emotional Compatibility Constraint:} Prevents scheduling emotionally demanding events during incompatible emotional states:
\[     \mathbf{C_4}: \textbf{If Emotion}_i = \textbf{angry or stressed},      \] 
\[     \quad \textbf{avoid scheduling sensitive or high-stakes meetings.}     \]
\end{itemize}

\subsection{Global Objective Function}

The optimal solution minimizes the following multi-objective function:
\begin{equation*}
\begin{split}
f(S, D) = & \alpha_1 f_{temporal}(S, D) \\
    &+ \alpha_2 f_{cognitive}(S, D) \\
    & + \alpha_3 f_{emotional}(S, D) \\
     &+ \alpha_i f_{j}(S, D)
\end{split}
\end{equation*}

where:
\begin{itemize}
    \item $S = \{s_1, s_2, \ldots, s_n\}$ represents a schedule (assignment of start times)
    \item $D = \{D_1, D_2, \ldots, D_n\}$ represents the duration of each tasks
    \item $f_{temporal}(S)$ measures temporal efficiency (e.g., minimizing idle time)
    \item $f_{cognitive}(S)$ evaluates cognitive load distribution
    \item $f_{emotional}(S)$ quantifies emotional well-being preservation
    \item $f_{j}(S)$ can be any arbitrary objective function complementing a specific business domain or environment
    \item $\alpha_1, \alpha_2, \alpha_3, \ldots, \alpha_i$ are weights reflecting the relative importance of each objective
\end{itemize}

This emotionally-informed CSP formulation enables calendars to adapt scheduling decisions based on real-time emotional states, creating a personalized, environment agnostic experience that balances productivity with user well-being.

\subsection{Emotion Detection}
\subsubsection{Facial recognition}
The most common and well-known method of emotion recognition is facial recognition which is reliably associated with specific facial behaviors \cite{ekman1971constants}. It has been observed that both humans and animals exhibit certain muscular movements linked to particular mental states \cite{FacialDL}. Image classification systems using various CNN-based models applied to large, high-resolution image datasets have achieved competitive accuracy, with the highest reaching 95\% \cite{krizhevsky2012imagenet,lv2014facial,correa2016emotion}. Another approach for facial emotion detection is through facial emotion recognition in videos. Some researchers have proposed a CNN-LSTM-based neural network, which was trained on the CREMA-D dataset and tested on the RAVDEES dataset for six basic emotions. Others enhanced the method by integrating a bi-modal user interface paradigm to complement the visual-facial modality with an additional audio-lingual modality \cite{VirvouTASK12}.

\subsubsection{Text-based emotion recognition}
\begin{figure}[h]
    \centering
    \includegraphics[width=0.5\textwidth]{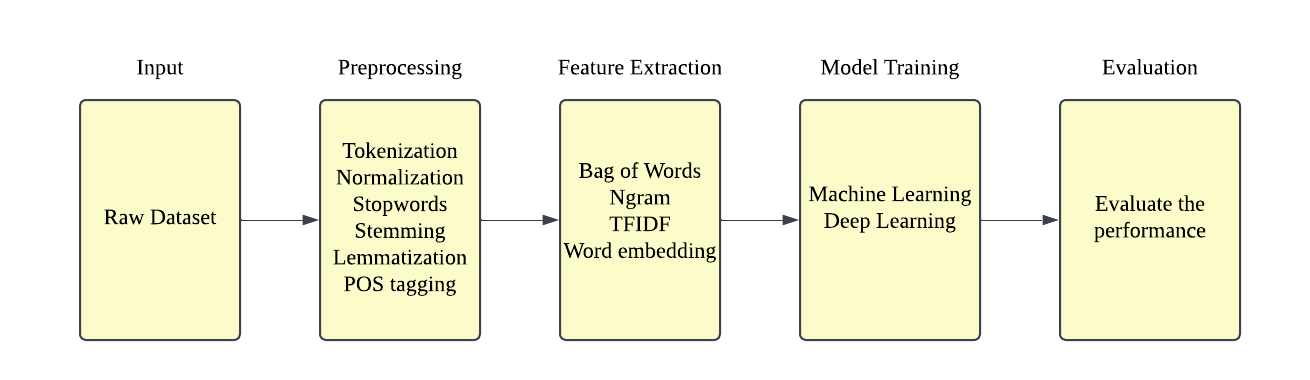}
    \caption{The process of text-based emotion recognition \cite{nandwani2021review}}
    \label{fig:text}
\end{figure}

Human language understanding in Natural Language Processing (NLP) research has become a significant area of study, with emotion detection being a crucial aspect. The process of text-based emotion recognition \cite{nandwani2021review} involves five key steps (Fig. \ref{fig:text}). First, a dataset is collected, typically consisting of text data. Commonly used and well-formatted datasets include SemEval, Stanford Sentiment Treebank (SST), and the International Survey of Emotional Antecedents and Reactions (ISEAR). The next step is preprocessing, which involves tokenization, normalization, removal of stopwords, part-of-speech tagging, stemming, and lemmatization to prepare the text for analysis. Feature extraction follows, utilizing techniques such as bag of words, n-grams, TF-IDF, and word embeddings to convert the text into a format suitable for modelling. During the model development stage, machine learning algorithms, such as K-Nearest Neighbors (KNN) and Naive Bayes \cite{suhasini2020emotion}, or deep learning models, including Bi-LSTM \cite{ragheb2019attention} are employed for emotion classification. Additionally, hybrid models that integrate both machine learning and deep learning approaches, such as Convolutional Neural Networks (CNN) and Bi-GRU models \cite{bharti2022text} are also explored. The performance of the developed models is evaluated by comparing them with existing approaches, utilizing standard evaluation metrics such as accuracy, precision, and recall.

\subsubsection{Device activity}
Using computer activity data, such as keyboard and mouse interactions, is a non-invasive, low-cost method for emotion detection. It is possible to apply different machine learning models, including decision trees, k-nearest neighbors (KNN), naive Bayes, AdaBoost, rotation forests, and Bayesian networks, to analyze keyboard activity \cite{Agata}. Rather than developing a single classifier for all participants and predefined emotional states, multiple binary classifiers were utilized to classify each emotion. The best results were achieved for fear and anger, with average performance for sadness, happiness, and boredom, and the poorest performance for surprise and disgust. Other researchers \cite{Lali} employed 50-fold cross-validation and applied KNN, multilayer perceptron (MLP), and support vector machines (SVM) as classifiers on self-collected mouse trajectory data to detect pleasant and unpleasant emotions. The KNN method (with k=5) and a 50\% decay time yielded the best performance. The multimodal user interface aspect can also be applied with keyboard activity as a modality and tailored for mobile devices \cite{AlepisV12}.

\subsubsection{Signal approach}
A prevalent approach for emotion detection utilizes physiological signals, particularly electroencephalogram (EEG) signals, to recognize emotions by analyzing three primary influential features\cite{bos2006eeg}:
\begin{itemize}
    \item Valence: Positive, happy emotions are associated with higher frontal coherence in alpha waves and increased beta power in the right parietal lobe, in contrast to negative emotions.
    \item Arousal: Excitement is characterized by higher beta power and coherence in the parietal lobe, accompanied by lower alpha activity.
    \item Dominance: The strength of emotion is typically represented in the EEG by an increased beta-to-alpha activity ratio in the frontal lobe, along with heightened beta activity in the parietal lobe.
\end{itemize}
Some researchers\cite{choppin2000eeg} have analyzed EEG signals and employed neural networks to classify six emotions based on emotional valence and arousal. Others\cite{Deepher} have utilized deep neural networks, such as LSTM, which have now achieved high prediction accuracy, reaching 99.9\%.

\subsection{Deep Learning Models}
\subsubsection{LSTM model}
\begin{figure}
    \centering
    \includegraphics[width=0.3\textwidth]{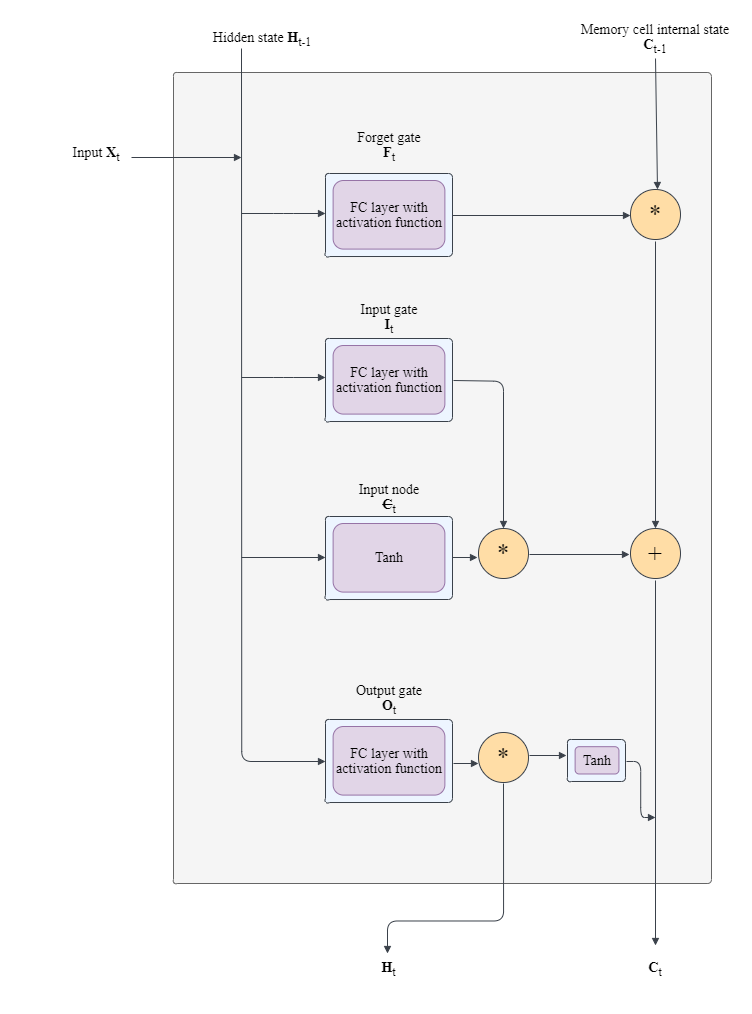}
    \caption{The architecture of LSTM model}
    \label{fig:LSTM}
\end{figure}

A recurrent neural network (RNN) is a type of deep learning model designed to process sequential data, such as time series. However, in practice, RNNs often struggle to learn long-term dependencies in data due to issues like the vanishing gradient problem.
LSTM networks \cite{hochreiter1997long} are a specialized variant of RNNs that address this limitation by effectively learning long-term dependencies. The architecture of an LSTM is depicted in Fig. \ref{fig:LSTM}. At each time step $t$, the LSTM receives the input $X_t$ and the previous hidden state $h_{t-1}$. It consists of three primary gates: the forget gate, which determines which information from the previous cell state should be discarded; the input gate, which controls the incorporation of new information into the cell state; and the output gate, which regulates the information output from the cell state as the current hidden state $h_t$. The cell state functions as the LSTM's long-term memory, while the hidden state captures short-term memory. These gates use sigmoid activation functions to manage the flow of information, and the tanh activation function is applied to scale the values in both the cell and hidden states. 

\subsubsection{GRU model}
\begin{figure}[h]
    \centering
    \includegraphics[width=0.35\textwidth]{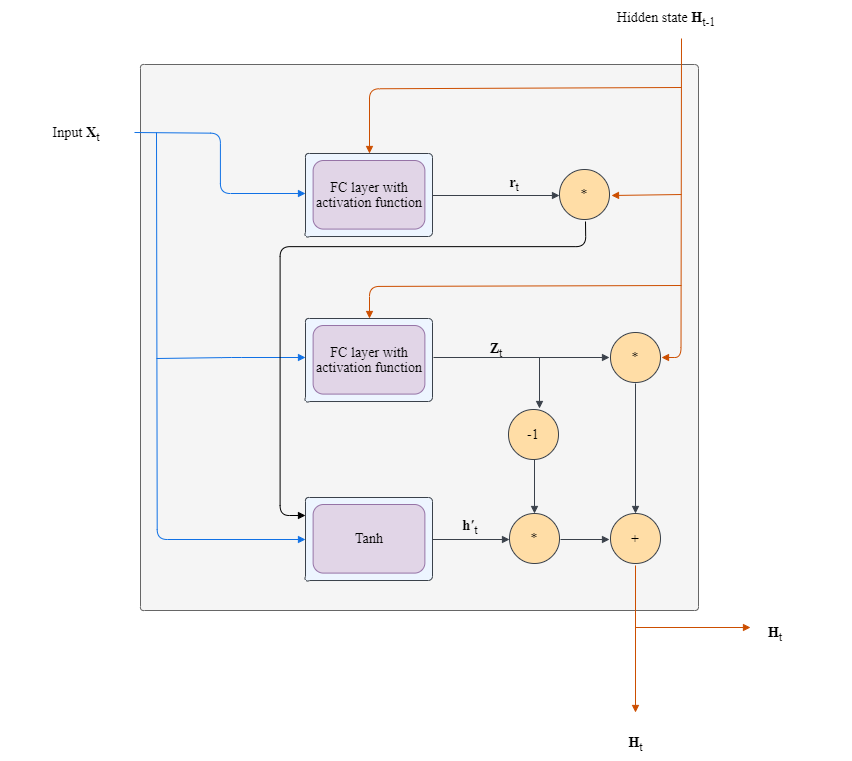}
    \caption{The architecture of GRU model}
    \label{fig:GRU}
\end{figure}

The GRU \cite{GRU} is a type of RNN architecture, closely related to the LSTM network. GRU is designed to model sequential data by enabling the selective retention or discarding of information over time. Compared to LSTM, GRU has a more streamlined architecture with fewer parameters, which often leads to improved training efficiency and reduced computational complexity.The GRU architecture, as depicted in Fig. \ref{fig:GRU}, consists of several key components designed to control the flow of information across time steps. It uses two gates: the reset gate ($r_t$) and the update gate ($z_t$). The reset gate determines how much of the previous hidden state should be discarded when calculating the candidate's hidden state, while the update gate controls how much of the previous hidden state should be retained. The candidate hidden state ($h_t'$) is calculated using the input at the current time step ($x_t$) and the reset gate, and is passed through a $\tanh$ activation function. The final hidden state ($h_t$) is a weighted sum of the previous hidden state and the candidate hidden state, with the update gate controlling the contribution of each. 

\section{Proposed Approaches}
\label{Sec:Proposed}

Two distinct approaches were employed for emotion detection in the context of UI/UX interface usage for calendar (time/event) optimization: biometric analysis utilizing ECG signals and heart rate data to predict the three emotion dimensions (Valence, Arousal, and Dominance), and computer activity analysis for emotion detection.

\subsection{Dataset}
\subsubsection{DREAMER \cite{Dreamer} Dataset for biometric signals analysis}
The DREAMER dataset consists of physiological recordings and ratings collected from 23 volunteers during an experiment in which they viewed 18 movie clips. The clips were selected and rated by Gabert-Quillen et al., and EEG and ECG signals were recorded during the experiment. Each participant provided self-reported mood ratings based on perceived arousal, potency, and dominance, using a 5-point scale.

The dataset includes two main variables: "stimuli" and "baseline." The "stimuli" variable contains data corresponding to the 18 movie clips, while the "baseline" variable includes data from neutral clips shown prior to each movie clip. The records for each movie clip are denoted as baseline{i} and stimuli{i}, representing the data for the $i^{th}$ movie clip.

For the ECG recordings, each entry is structured as an M×2 matrix, where M represents the number of samples, and each column contains samples from the two ECG channels.

\subsubsection{DUX \cite{DUX} Dataset for computer activity analysis}
This dataset captures data from user interactions with the keyboard and mouse, alongside emotion data collected through the iMotions facial coding module and manual annotation. The dataset encompasses consistent data across all 36 test sessions with triggers enabled. Each file stores event data in a row-wise format, including event types (e.g., key presses, mouse clicks), target UI elements, modifier key states, and mouse coordinates. Additionally, the dataset features 12 mood intensities automatically detected by the iMotions system, as well as corresponding manually annotated mood data. These emotional intensities range from 0\% to 100\%.

\subsection{Methodology}
\subsubsection{Biometric signals}
\begin{figure}[h]
    \centering
    \includegraphics[width=0.34\textwidth]{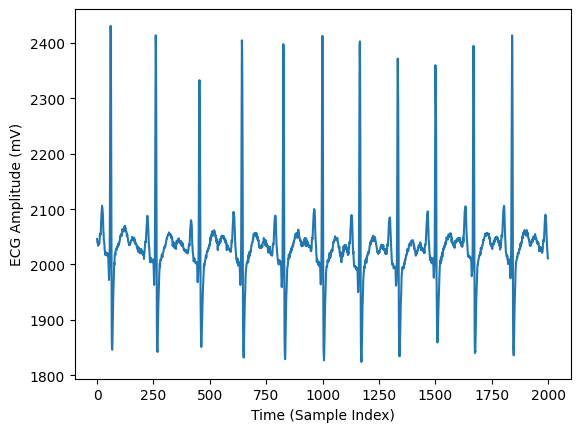}
    \caption{An example of ECG stimuli signal}
    \label{fig:ecg}
\end{figure}

\begin{figure}[h]
    \centering
    \includegraphics[width=0.34\textwidth]{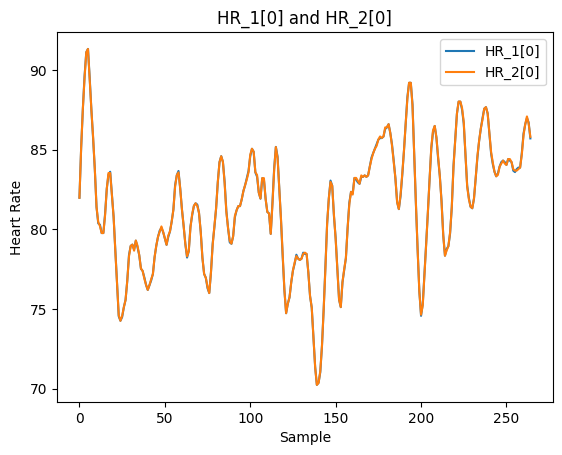}
    \caption{Heart rate signals from both ECG channels (HR\_1 and HR\_2).}
    \label{fig:hr_plot}
\end{figure}

For data preparation, we retrieved the "stimuli" ECG data along with three dimensions of emotion measurement (Valence, Arousal, and Dominance) for 23 participants. Each participant's dataset contained 18 records corresponding to 18 clips from the original raw signal. The heart rate signal was extracted from the ECG data Fig. \ref{fig:ecg} by identifying the R-peak and calculating the R-R interval in seconds, from which the HR in beats per minute was derived. Since two ECG channels were available, we plotted the HR alongside the corresponding ECG signals from both channels (HR\_1 and HR\_2), as shown in Fig. \ref{fig:hr_plot}. The results indicated that the HR signals from both channels were identical. Therefore, for the training input, we selected HR\_1. Next, we performed signal normalization and applied zero-padding to ensure all signals had the same length. Finally, the 5-point scale for each emotion dimension was transformed into two categories: low and high. Given that the data is time series, input sequences of a specified window size were generated from the heart rate data and corresponding labels, where each sequence represents a window of time steps and the label corresponds to the next emotion value. The three distinct models were then trained using a LSTM architecture with a dropout rate of 0.5 to mitigate overfitting due to the small dataset,and three separate models were trained utilizing GRU. The target labels corresponded to the three dimensions of emotion measurement.

\subsubsection{Computer Activity}

\begin{figure}[h]
    \centering
    \includegraphics[width=0.4\textwidth]{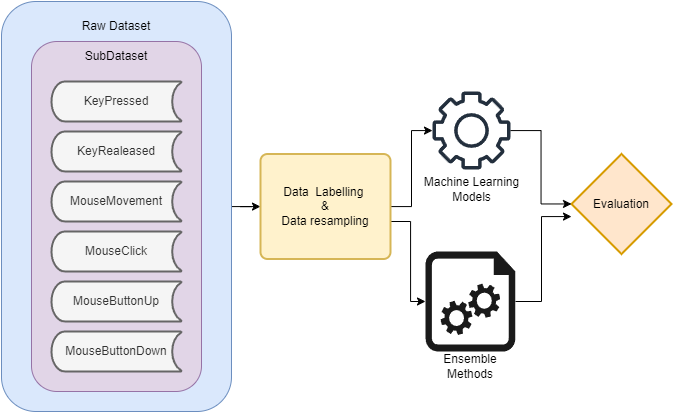}
    \caption{Computer activity workflow}
    \label{fig:computer_activity_workflow}
\end{figure}

Fig. \ref{fig:computer_activity_workflow} represents the workflow for using computer activity to predict emotion. For data preparation, the dataset was partitioned into six sub-datasets based on distinct computer activities: four related to mouse actions (MouseMovement, MouseClick, MouseButtonUp, and MouseButtonDown) and two related to keyboard activities (KeyPressed and KeyReleased). Each sub-dataset was then labeled with the ground truth emotion value, which was determined by selecting the highest emotion score among the 12 mood intensities automatically detected by the iMotions system.

\paragraph{MouseMovement Sub-dataset} 
The input values for this sub-dataset consist of the x and y coordinates of the mouse cursor within the application at the time of the event. The predicted value is the emotion type associated with the highest intensity score detected.

\paragraph{MouseClick, MouseButtonUp, and MouseButtonDown Sub-datasets}
In these sub-datasets, the input values include the mouse button indicator (denoting which mouse button is pressed or released), along with the x and y coordinates of the mouse cursor in the application at the time of the event. The predicted value is the emotion type, as detected by the iMotions system.

\paragraph{KeyPressed and KeyReleased Sub-datasets} 
These sub-datasets include several input features for keyboard events:
\begin{itemize}
    \item \textbf{alt, control, shift, meta}: Boolean values indicating whether the respective modifier keys were pressed during the event.
    \item \textbf{key}: The key being pressed or released, with personal information replaced by ``ANONYMIZED'' where necessary.
    \item \textbf{repeat}: A boolean value indicating whether the event was repeated due to the key being held down (FALSE for the first emission, TRUE for subsequent emissions).
\end{itemize}
The predicted value is the emotion type detected at the time of the event.

Due to the imbalance in the data distribution across the different sub-datasets, we applied the Synthetic Minority Over-sampling Technique (SMOTE) \cite{AzureML_SMOTE} to address this issue. Specifically, the resampling was performed based on the number of records within each sub-dataset. For the MouseMovement and KeyReleased sub-datasets, we resampled to 5,000 records for each emotion category. For the MouseClick, KeyPressed, MouseButtonUp, and MouseButtonDown sub-datasets, we resampled to 500 records per emotion category. Subsequently, several machine learning models were applied to evaluate and compare their performance across the different sub-datasets. These models included a baseline logistic regression, random forest, SVM, decision tree, MLP, and Naive Bayes. Additionally, ensemble methods, such as gradient boosting machine, XGBoost, AdaBoost, and BaggingClassifier, were employed.

\section{Experimental Evaluation}
\label{Sec:Experimental}

\subsection{Biometric Analysis Results} 
To evaluate the performance of using heart rate signals for detecting levels of three emotion dimensions, cross-entropy (CE) loss  (Eq. \ref{eq:CE}) was employed as the loss function, and accuracy (Eq. \ref{eq:Acc}), based on correct predictions, was used to measure model performance.
\begin{equation}
\text{Cross-Entropy Loss} = - \sum_{i=1}^{N} y_i \log(p_i)
\label{eq:CE}
\end{equation}
where:
\begin{itemize}
    \item \( N \) denotes the total number of classes.
    \item \( y_i \) is the true label for class \( i \) (1 if the class corresponds to the true label, 0 otherwise).
    \item \( p_i \) represents the predicted probability for class \( i \), derived from the model's output.
\end{itemize}

\begin{equation}
    \text{Accuracy} = \frac{\text{Number of Correct Predictions}}{\text{Total Number of Predictions}}
    \label{eq:Acc}
\end{equation}
\\

Table \ref{bio_mat} presents the performance of the LSTM and GRU models for the three emotional measurement dimensions. The test accuracy for the three emotion dimensions varies across the different architectures post-training. For the LSTM model, the accuracy for Valence (75\%) is slightly higher than for Arousal (71.88\%) and significantly higher than for Dominance (62.5\%). The CE loss for all dimensions is below 0.025, with Arousal exhibiting the highest CE loss (0.0230), which is similar to Dominance's CE loss (0.0219), while Valence shows the lowest CE loss (0.0164).

For the GRU model, Valence achieves the highest accuracy (84.88\%) and the lowest CE loss (0.0143) among all results. This is followed by Arousal, with an accuracy of 71.88\% and a loss of 0.0187. The Dominance label shows the lowest accuracy (68.75\%) and a loss of 0.0197.

Based on these performance metrics, it can be concluded that both the LSTM and GRU models effectively predict the two levels (low and high) of the three emotion dimensions. However, the GRU model outperforms the LSTM model, exhibiting higher accuracy and lower loss for the same labels. Additionally, the performance of the Dominance label is the least favorable, which can be attributed to the fact that Dominance is more closely related to electrodermal activity and electromyography, while HR may not be a major contributing factor. The slightly lower performance of the Arousal label, compared to Valence, may be due to the nature of the data. Arousal is more pronounced in dynamic situations (e.g., during exercise), whereas in static contexts (such as the experiment involving the viewing of 18 movie clips), it is less noticeable. Furthermore, the variability in the baseline heart rate among individuals may reduce the model's generalization ability.

\begin{table}[h!]
\centering
\caption{Accuracy \& Cross Entropy (CE) Loss of LSTM and GRU models }

\begin{tabular}{|c|c|c|c|c|}
\hline
\multirow{2}{*}{\textbf{Category}} & \multicolumn{2}{c|}{\textbf{LSTM}} & \multicolumn{2}{c|}{\textbf{GRU}} \\ 

\cline{2-5}
&Accuracy & CE Loss & Accuracy & CE Loss \\ 

\hline
\textbf{Valence} & \underline{75\%} & \underline{0.0164} & \underline{\textbf{84.38\%}} & \underline{\textbf{0.0143}} \\ \hline
\textbf{Dominance} &62.5\% &0.0219 & 68.75\%&0.0197 \\ \hline
\textbf{Arousal} & 71.88\%&0.0230 & 71.88\%& 0.0187\\ \hline
\end{tabular}
\label{bio_mat}
\end{table}

\begin{table}[h!]
\centering
\caption{Mouse Movement/Click Performance (Accuracy)}                  
\begin{tabular}{|c|c|c|}
\hline
\textbf{Method}         & \textbf{Movement (Acc.)}      & \textbf{Click (Acc.)} \\ \hline
Logistic Regression     & 24.25\%                       & 57.14\%               \\ \hline
Random Forest           & \textbf{86.59\%}              & 93.57\%               \\ \hline
Support Vector Machine  & 51.75\%                       & 64.42\%               \\ \hline
XGBoost                 & 84.35\%                       & 93.57\%               \\ \hline
LightGBM                & 83.2\%                        & 92.86\%               \\ \hline
Decision Tree           & 85.15\%                       & \textbf{93.86\%}      \\ \hline
AdaBoost                & 30.66\%                       & 51.29\%               \\ \hline
Multi-layer Perceptrons & 48.09\%                       & 63.29\%               \\ \hline
Naive Bayes             & 24.23\%                       & 53.43\%               \\ \hline     
\end{tabular}
\label{tab:mouse_movement_click}
\end{table}

\begin{table}[h!]
\centering
\caption{Key Pressed/Released Performance (Accuracy)}
\begin{tabular}{|c|c|c|}
\hline
\textbf{Method}             & \textbf{Pressed (Acc.)}   & \textbf{Released (Acc.)}  \\ \hline
Logistic Regression         & 31.5\%                   & 24.28\%                   \\ \hline
Random Forest               & 52.625\%                  & 96.28\%                   \\ \hline
Support Vector Machine      & 35.62\%                   & \textbf{96.31\%}                   \\ \hline
XGBoost                     & 52.25\%                  & \textbf{96.31\%}                   \\ \hline
LightGBM                    & 51.625\%                  & 95.75\%                   \\ \hline
Decision Tree               & 52.625\%                  & 96.25\%                   \\ \hline
AdaBoost                    & 30.375\%                  & \textbf{96.31\%}                   \\ \hline
Multi-layer Perceptrons     & 32.5\%                   & \textbf{96.31\%}                  \\ \hline
Bagging classifier          & \textbf{53\%}             & 96.28\%                  \\ \hline
\end{tabular}
\label{tab:key_pressed_released}
\end{table}

\begin{table}[h!]
\centering
\caption{Button Up/Down Performance (Accuracy)}
\begin{tabular}{|c|c|c|}
\hline
\textbf{Method} & \textbf{Up (Acc.)} & \textbf{Down (Acc.)} \\ \hline
Logistic Regression     & 49.57\%               & 53\%              \\ \hline
Random Forest           & 93.43\%               & \textbf{93.71\%}  \\ \hline
Support Vector Machine  & 64.57\%               & 65.14\%           \\ \hline
XGBoost                 & 93.43\%              & \textbf{93.71\%} \\ \hline
LightGBM                & \textbf{93.86\%}     & 89.71\%           \\ \hline
Decision Tree           & 93.71\%               & 92.14\%           \\ \hline
AdaBoost                & 61\%                  & 51.71\%           \\ \hline
Multi-layer Perceptrons & 64.51\%               & 61\%              \\ \hline
Naive Bayes             & 56.29\%               & 48.57\%           \\ \hline    
\end{tabular}
\label{tab:button_up_down}
\end{table}

\subsection{Computer Activity Results}
The accuracy (Eq. \ref{eq:Acc}) was used as the primary metric to evaluate the performance of the machine learning models and ensemble methods for each sub-dataset.
Table \ref{tab:mouse_movement_click} presents the performance of various machine learning models and ensemble methods applied to the Mouse Movement activity, with the Random Forest model achieving the highest accuracy of 86.59\%. Table \ref{tab:mouse_movement_click} displays the performance for the Mouse Click activity, where the Decision Tree model outperforms others with an accuracy of 93.86\%. Table \ref{tab:key_pressed_released} shows the results for the Key Pressed activity, with the ensemble method Bagging Classifier attaining the best performance at 53\%. Table \ref{tab:key_pressed_released} highlights the performance of the Key Released activity, where the Support Vector Machine (SVM), XGBoost, AdaBoost, and Multi-layer Perceptron models all achieve the highest accuracy of 96.31\%. Table \ref{tab:button_up_down} presents the performance for the Mouse Button Up activity, with the ensemble method LightGBM achieving the highest accuracy of 93.86\%. Finally, Table \ref{tab:button_up_down} shows the results for the Mouse Button Down activity, where Random Forest and XGBoost models both achieve the highest accuracy of 93.17\%.

Across all the tables, it can be observed that among the machine learning models, Random Forest performs the best, while XGBoost emerges as the most effective ensemble method. However, the performance of the machine learning models generally surpasses that of the ensemble methods.

Among all the computer activity sub-datasets, mouse-related activities consistently demonstrated stable and high performance in emotion recognition, with accuracy levels generally around 90\%. This stability suggests that mouse movements and interactions are strong indicators for detecting emotional states. In contrast, the performance of keyboard-related activities was more variable. Specifically, the Key Released activity achieved the highest accuracy 96.31\%, while the Key Pressed activity exhibited significantly lower accuracy. One possible explanation for this discrepancy may be the quality of the Key Pressed sub-dataset, which could influence the model's ability to accurately detect emotions from keyboard interactions.

\section{Conclusion}
\label{Sec:Conclusion}

This study presents and evaluates two complementary approaches for emotion detection within the novel context of calendar optimization systems. Our extensive comparative analysis yields several significant insights with implications for emotion-aware interface design.
Future research will pursue an extension of our analysis to mobile platforms by incorporating touch-based interaction metrics and developing specialized models for touchscreen dynamics.
Moreover, we will enhance the biometric approach by investigating multimodal fusion techniques and we aim to develop and evaluate a fully integrated emotion-aware calendar optimization system in real-world settings.

\bibliographystyle{IEEEtran}
\bibliography{references}

\end{document}